\def\year{2019}\relax
\documentclass[letterpaper]{article} 
\usepackage{aaai19}  
\usepackage{times}  
\usepackage{helvet} 
\usepackage{courier}  
\usepackage[hyphens]{url}  
\usepackage{graphicx} 
\usepackage{graphicx}  
\usepackage{amsmath}\usepackage{subcaption}
\usepackage{amsfonts}
\usepackage{amsmath,amsthm,amssymb}
\usepackage[ruled ]{algorithm2e}

\DeclareMathOperator*{\argmax}{arg\,max}
\DeclareMathOperator*{\argmin}{arg\,min}

\newtheorem{theorem}{Theorem}
\newtheorem{proposition}[theorem]{Proposition}
\newtheorem{lemma}[theorem]{Lemma}

\newcommand{\citet}[1]{\citeauthor{#1}~\shortcite{#1}}
\newcommand{\citep}{\cite}

\begin{document}
%
\title{Large Scale Learning of Agent Rationality in Two-Player Zero-Sum Games
}
\author{Chun Kai Ling,\textsuperscript{\rm 1}
Fei Fang,\textsuperscript{\rm 1}
J. Zico Kolter\textsuperscript{\rm 1, \rm 2}\\
\textsuperscript{\rm 1} School of Computer Science, Carnegie Mellon University\\
\textsuperscript{\rm 2} Bosch Center for Artificial Intelligence\\
\{chunkail, feif, zkolter\}@cs.cmu.edu
}
\maketitle
\begin{abstract}
With the recent advances in solving large, zero-sum extensive form games, there is a growing interest in the inverse problem of inferring underlying game parameters given only access to agent actions. Although a recent work provides a powerful differentiable end-to-end learning frameworks which embed a game solver within a deep-learning framework, allowing unknown game parameters to be learned via backpropagation, this framework faces significant limitations when applied to boundedly rational human agents and large scale problems, leading to poor practicality. In this paper, we address these limitations and propose a framework that is applicable for more practical settings. 
First, seeking to learn the rationality of human agents in complex two-player zero-sum games, we draw upon well-known ideas in decision theory to obtain a concise and interpretable agent behavior model, and derive solvers and gradients for end-to-end learning. 
Second, to scale up to large, real-world scenarios, we propose an efficient first-order primal-dual method which exploits the structure of extensive-form games, yielding significantly faster computation for both game solving and gradient computation. 
When tested on randomly generated games, we report speedups of orders of magnitude over previous approaches. We also demonstrate the effectiveness of our model on both real-world one-player settings and synthetic data.
\end{abstract}
\section{Introduction}
Game theory has traditionally been centered around finding players' strategies in equilibrium.
In recent years, there has been growing interest in the inverse setting of learning game parameters from observed player actions \cite{vorobeychik2007learning,blum2014learning,waugh2011computational}. 
Recent work by \citet{ling2018game} tackle this problem in the zero-sum setting by providing an end-to-end learning framework to learn game parameters such as payoff matrices and chance node probability distributions assuming the actions are sampled from the Quantal Response equilibrium. At the core of the framework is a differentiable game solving module.

However, their proposed method suffers from two major flaws. Firstly, the assumption that players behave in accordance to the QRE severely limits the space of player strategies, and is known to exhibit pathological behavior even in one-player settings. Second, their solvers are computationally inefficient and are unable to scale.

Our work addresses these deficiencies in two ways. First, we propose the Nested Logit Quantal Response equilibrium (NLQRE), which draws upon ideas from
from behavioral science and allows for varying levels of player rationality at each stage of the game. We show that the NLQRE is strictly more general than the models considered by \citeauthor{ling2018game}, and may not be replicated by a straightforward scaling of payoff matrices. We derive the required gradients and show that player rationality can be learned via gradient descent can be learned using the same end-to-end learning framework. 
Second, we substantially reduce training time by reformulating the backward pass as a min-max convex optimization problem and uses state-of-the-art first order primal-dual methods for both the forward pass and backward pass. 
Unlike previous work, which relied on second-order methods, our first-order solver does not require explicit formation of Hessians and only requires access to a fast best-response oracle. In our evaluation with random payoff matrices and one-card poker, we report orders of magnitude of speedups. Lastly, we evaluate the NLQRE on real-world data in an one-player information gathering game and provide qualitative insights.  In total, we believe that our work is a significant step towards the practical learning of human behavior in zero-sum settings. 

\section{Background and related work}
\begin{figure}
    \centering
    \includegraphics[width=0.46\textwidth]{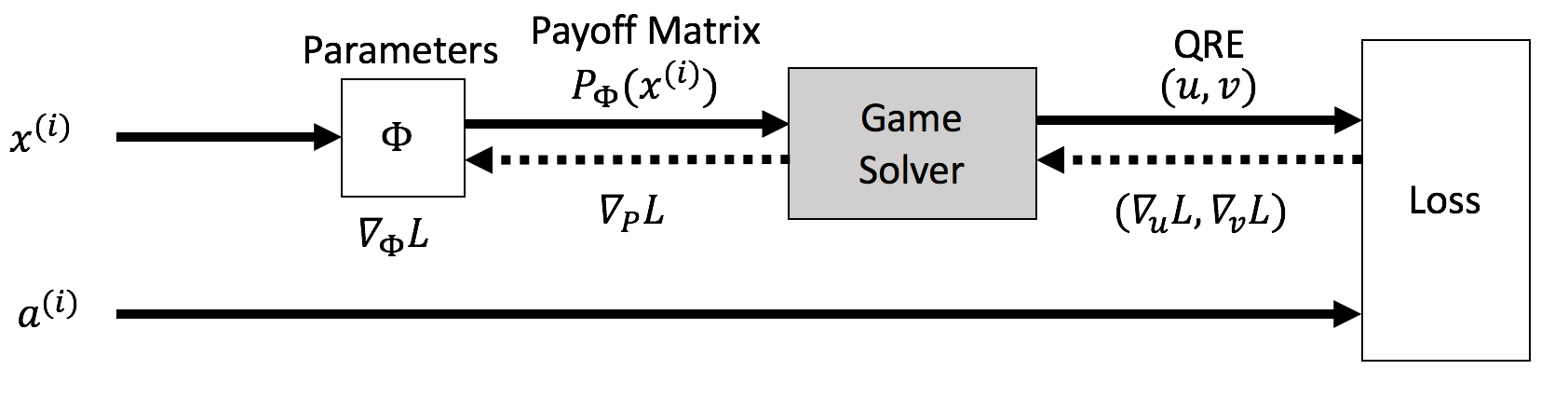}
    \caption{The framework used by \citeauthor{ling2018game}}
    \label{fig:overall_arch}
\end{figure}

Although much less well studied than traditional equilibrium finding, there are several approaches aimed at the task of learning games in the setting where underlying game payoffs are unknown. These include methods which rely on specific game structure such as symmetry \cite{vorobeychik2007learning}, operate in an active setting \cite{blum2014learning}, or focus primarily on normal form games and straightforward linear settings \cite{waugh2011computational}. 
\citet{ling2018game} provide an alternative framework which embeds a differentiable game solver within another gradient based learner (e.g., a deep network), as illustrated in Figure~\ref{fig:overall_arch}.
This enables game parameters to be learned via simple gradient descent. We now describe their framework briefly. 

Suppose $P_{\Phi}(x)$ is the zero-sum payoff matrix given some features $x$ and game parameters $\Phi$ which we wish to learn. The game solver takes in $P_{\Phi}(x)$ and outputs the QRE $(u, v)$, which correspond to mixed strategies of the min and max player. During training, the log loss $L(a, u, v)$ of the solver's predicted strategies is computed against observed actions $a$. The game parameters $\Phi$ are then optimized by minimizing $L$. This is performed by propagating gradients backwards through the game solver and performing gradient descent, where the required gradients for the backward pass are readily derived by using the implicit function theorem. 
The training phase is summarized in Algorithm~\ref{alg:lck_prev_work}.
We will now touch on two key ideas from decision and game theory, which will eventually culminate in the proposed NLQRE. 

\begin{algorithm}
\SetAlgoLined
 \KwIn{training data $\{(x^{(i)}, a^{(i)})\}$, learning rate $\eta$, $\Phi_\text{init}$}
 \For{ep in $\{0, \dots, \text{ep}_\text{max} \}$}
 {
  Sample $(x^{(i)}, a^{(i)})$ from training data\;
  Forward pass: Compute $P_\Phi(x^{(i)})$, QRE $(u, v)$ and loss $L(a^{(i)}, u, v)$\;
  Backward pass: Compute gradients $\nabla_u L, \nabla_v L, \nabla_P L, \nabla_\Phi L$\;
  Update parameters: $\Phi \leftarrow \Phi - \eta \nabla_\Phi L$\;
 }
 \caption{Learning game parameters $\Phi$ using SGD}
 \label{alg:lck_prev_work}
\end{algorithm}
\subsection{Nested-logit choice models}
One of the fundamental research problems in behavioral science is to mathematically model seemingly irrational (or non-utility maximizing) human behavior. Among the most important models is the class of the random utility models (RUM) \cite{thurstone1927law}.
The Logit model\footnote{Logits are more commonly known by the machine learning community as the `softmax' operator.} is the most notable RUM, where 
given a set of alternatives $\mathcal{A}$ each with (known) utility $U_a$, the probability that alternative $a$ is picked is $u^*_a = \frac{\exp(U_a)}{\sum_{a'\in\mathcal{A}} \exp(U_{a'})}$. It is equivalent to the probability that alternative $a$ has highest utility under Gumbel noise, i.e., $u^*_a = \mathbb{P} \left(\argmax_{a'} (U_{a'} + \epsilon_{a'}) = a \right)$, where $\epsilon_{a'}$ are  i.i.d. Gumbel distributed.

However, the logit model suffers from limitations. This includes classic `red and blue bus' pathologies\footnote{Suppose there are 3 alternatives for transport -- a red bus, a blue bus, and a car. The player derives the same utility for each alternative, $x_\text{car}=x_\text{red}=x_\text{blue}$. Applying the logit model gives an equal probability of choosing each vehicle. One would however, expect the car to be taken with probability $1/2$ and each bus to be chosen with probability $1/4$, since the color of buses should have no impact on decisions.} \cite{luce2012individual} which restrict the class of behaviors permitted. Specifically, logit models obey the property of \textit{independence of irrelevant alternatives}, which does not take into account cases when alternatives are `qualitatively' similar.

\textit{Nested-logit} (NL) models \cite{train2009discrete} address this limitation by grouping fundamentally similar alternatives together and allows for correlations between $\epsilon$'s belonging to the same group.
In a two-level NL model, $\mathcal{A}$ is divided into $K$ disjoint clusters, with   alternatives $a$ belonging to cluster $k(a)$ chosen with probability
{
\begin{align*}
    u^*_a &\propto \exp\left(U_a/\lambda_{k(a)}\right) \left( \sum_{a' \in k(a)} \exp\left(U_{a'}/\lambda_{k(a)}\right) \right)^{\lambda_{k(a)}-1},
\end{align*}
}\normalsize
where $\lambda$'s are parameters governing noise correlation. These probabilities may be interpreted as a two-stage decision making process: in the first stage, a cluster is chosen, and in the second stage, the specific action is selected based on (scaled) softmax on $U_a$ within the cluster. The probability of choosing each cluster in the first stage is given by the softmax over the (scaled) log-sum-exp of each cluster. 
When $\lambda=1$, the standard logit model is recovered, and when $\lambda \rightarrow 0^+$, the `elimination by aspects' is obtained \cite{tversky1972elimination}. NL models can have multiple layers, leading to a NL tree representing the nested grouping. The reader is directed to the book by \citeauthor{train2009discrete} for background about nested logits and their various interpretations. 

\subsection{Quantal response equilibria (QRE)}
We now turn our attention to 2-player games. Seminal work by McKelvey propose QRE 
as a noisy alternative to NE. \cite{mckelvey1995quantal}. Similar to logit choice models, the QRE is the equilibrium obtained when payoffs are perturbed by noise obeying a Gumbel distribution. 
Formally, $(u^*,v^*)$ is a QRE of a normal form game with action sets $\mathcal{A}_\textrm{u}$ and $\mathcal{A}_\textrm{v}$ for the two players and payoff matrix $P$ if 
\begin{align*}
u^*_a &= \frac{\exp(-Pv^*/\lambda)_a}{\sum\limits_{a' \in \mathcal{A}_\textrm{u}}\exp(-Pv^*/\lambda)_{a'}},v^*_a= \frac{\exp(P^Tu^*/\lambda)_a}{\sum\limits_{a' \in \mathcal{A}_\textrm{u}}\exp(P^Tu^*/\lambda)_{a'}},
\end{align*}
%
where $1/\lambda$ is a parameter governing the level of agent rationality. 
Observe that as $\lambda \rightarrow \infty$, players behave uniformly at random, while $(u^*, v^*)$ approaches a NE as $\lambda \rightarrow 0^+$. 
For zero-sum games, it is further known \cite{mertikopoulos2016learning} that QRE is the unique solution of the following convex-concave program
\begin{align*}
    \min_{u \in \mathbb{R}^n} 
    \max_{v \in \mathbb{R}^m} & \quad
    u^T P v + \lambda \sum_{a} u_a \log u_a - \lambda \sum_{a} v_a \log v_a \nonumber\\
    \text{subject to} & \quad 1^Tu = 1, \quad 1^Tv = 1.
\end{align*}
%

For a two-player extensive form game characterized by a game tree with information sets $\mathcal{I}_\textrm{u}$ and $\mathcal{I}_\textrm{v}$ for the min and max player respectively, \citeauthor{ling2018game} show that when $\lambda=1$, the QRE in reduced normal form of the game is equivalent to the solution of the following regularized min-max problem, where $u$ and $v$ are the players' strategies in \textit{sequence form} \cite{von1996efficient}.
\begin{align}
\min_{u} 
    \max_{v} u^T P v + \sum_{h \in \mathcal{I}_u} \sum_{a \in \mathcal{A}_h} u_a \log \frac{u_a}{u_{p_h}} \nonumber \\
    - \sum_{h \in \mathcal{I}_v} \sum_{a \in \mathcal{A}_h} v_a \log \frac{v_a}{v_{p_h}} \nonumber \\
    \text{subject to} \quad Eu = e, \quad Fv = f \label{eqn:seq_regularization}
\end{align}
In the above, $P$ is the sequence form payoff matrix and $E$ and $F$ are the sequence form constraint matrices. $\mathcal{A}_h$ denotes the possible actions at information set $h$, while $p_h$ is the action (from the same player) preceding $h$. In the sequence form, one works with \textit{realization plans} $u, v$ as opposed to probability vectors. These realization plans represent probabilities of choosing a given sequence, while the constraint matrices $E,F$ are matrices containing $\{0, 1, -1\}$ and contain parent-child relationships in the game tree. The sequence form is significantly more compact than the normal form while retaining virtually all of its strategic elements.

\section{Nested-logit quantal response equilibria}
Our proposed Nested-logit QRE (NLQRE) is a generalization of both the QRE (in zero-sum games) and NL models. That is, it generalizes NL models to two player zero-sum games, or equivalently, extends the QRE by permitting a more general nested logit structure. This allows us to model a far wider range of player behaviors, and in particular, cases where player rationality varies between stages of the game. We assume that the grouping of actions within each information set is known a-priori. 
The NLQRE is given by the unique solution to the following optimization problem
\begin{align}
&\min_{u} 
    \max_{v} \; & u^T P v + \sum_{h \in \mathcal{I}_\textrm{u}} \lambda_h \sum_{a \in \mathcal{A}_h} u_a \log \frac{u_a}{u_{p_h}}    \nonumber\\
    & & - \sum_{h \in \mathcal{I}_\textrm{v}} \lambda_h \sum_{a \in \mathcal{A}_h} v_a \log \frac{v_a}{v_{p_h}} \nonumber \\
    &\text{subject to} & \quad Eu = e, \quad Fv = f.
\label{eqn:seq_regularization_lambd}
\end{align}
%
The NL model is recovered in a one-player setting (i.e., $Pv$ is a constant vector) and the QRE is recovered when there is no nesting and $\lambda$'s are equal.
\citeauthor{ling2018game} assumes that all the $\lambda$'s are equal to $1$ and focus on learning entries of the payoff matrix $P$ by exploiting the smoothness of QRE solutions. This allows us to employ gradient-based approaches \cite{amin2016gradient} for learning. 
In this paper, we do not assume $\lambda$'s are known in our solution concept and instead treat them as parameters to be learned.

The additional representation power brought by introducing $\lambda$'s to \ref{eqn:seq_regularization} cannot be achieved by a simple scaling of the payoff matrix in the original formulation by \citeauthor{ling2018game}, even in the non-nested, simultaneous move normal form settings. To see why this is so, consider the game of symmetric rock-paper-scissors with non-uniform rewards (i.e., the payoffs for winners depend on their specific action). Suppose the game is played between `strong' and `weak' players, and this is reflected by low and high $\lambda$ parameters respectively. Due to differing $\lambda$'s for each player, the strategies of the two players in equilibrium are different. However, scaling $P$, or even changing individual payoffs for winners (while maintaining symmetry) can only result in symmetric equilibrium.

\textit{Remark 1}. Readers familiar with nested logits may recall that the most common form of nested logits do not admit chance nodes (or in our 2-player setting, parallel information sets). It may be shown that there is a natural way of doing so by considering representing each alternative as a pure strategy in the \textit{reduced normal form}, and by nesting each action based on the information sets which have a non-zero probability of being reached. The details are presented in the appendix.

\textit{Remark 2}. The expression in \eqref{eqn:seq_regularization_lambd} is fairly general. Broadly speaking, our framework allows for 2 types of nesting. First we allow for nesting via information sets (i.e., each information set gets its own $\lambda$, see Remark 1), and second, by clustering actions within an information set, which is achieved by introducing intermediate information sets (e.g., the `red and blue bus' example). Our experiments in Section~\ref{sec:expts} focus on the former. However \eqref{eqn:seq_regularization_lambd} and our proposed solver is able to handle the latter case, assuming that the nesting structure is known a-priori.

\textit{Remark 3}. The form of regularization is known as in \eqref{eqn:seq_regularization_lambd} is known as the \textit{dilated entropy regularization} and the $\lambda$'s may be interpreted as governing the degree of regularization or smoothing. Its form was first introduced by \citeauthor{hoda2010smoothing}, and follow-up work by \citeauthor{kroer2017theoretical} provided specific strong-convexity bounds on the regularizer. In particular, a specific instantiation of $\lambda$'s results in a best-response problem which is 1-strongly convex. The authors exploit this fact to yield some of the fastest solvers for Nash equilibria of two player zero-sum extensive form games. Note however, that their motivation is primarily computational in nature, as opposed to modelling, which is the case for us.
\subsection{NLQRE solver}
Following the ideas proposed by \citeauthor{ling2018game}, we present a naive solver for the NLQRE based on Newton's method. Denote $\mathcal{C}_a$ and $\mathcal{C}_{a'}$ as sets of possible information sets immediately following $a$ or $a'$. Define
$J_a=\sum_{h' \in C_a} \lambda_{h'}$, and let $\rho_a$ be the information set immediately preceding the action $a$, i.e. $h$ where $a \in \mathcal{A}_h$. The KKT conditions for \eqref{eqn:seq_regularization_lambd} are, for all $ h' \in \mathcal{I}_v, a' \in \mathcal{A}_{h'}$, and for all $h \in \mathcal{I}_u, a \in \mathcal{A}_h$,
    \\
\begin{align}
    (Pv)_a + \lambda_{\rho_a} ( 1 + \log \frac{u_a}{u_{p_h}}) - J_a + \sum_{c \in \mathcal{C}_a} \mu_c - \mu_{h}
    &= 0 \nonumber
    \\
    (P^Tu)_{a'} - \lambda_{\rho_{a'}} (1 + \log \frac{v_{a'}}{v_{p_{h'}}}) + J_{a'} 
    + \sum_{c \in \mathcal{C}_{a'}} \nu_c - \nu_{h'}
    &= 0  \nonumber \\ 
 	Eu-e = 0 \qquad 
	Fv-f = 0. \qquad \qquad & \label{eq:old_newton_forward}
\end{align}
These are necessary and sufficient conditions for NLQRE, implying that the NLQRE can be found by applying Newton's method to (\ref{eq:old_newton_forward}), yielding the following updates 
\begin{gather}
    \begin{bmatrix}
    \Xi(u) & P & E & 0 \\
    P^T & -\Xi(v) & 0 & F \\
    E^T & 0 & 0 & 0 \\
    0 & F^T & 0 & 0
    \end{bmatrix}
    \begin{bmatrix}
    \Delta u \\ \Delta v \\ \Delta \mu \\ \Delta \nu
    \end{bmatrix}
    = -g(u, v, \mu, \nu), \\
    \text{where} \quad  \Xi(u)_{ab} = 
    \begin{cases}
    \frac{\lambda_{\rho_a}+ \sum_{h' \in \mathcal{C}_a} \lambda_{h'}}{u_a}, \enskip a=b\\
    -\frac{\lambda_{\rho_a}}{u_b}, \enskip p_{\rho_a}=b\\
    -\frac{\lambda_{\rho_b}}{u_a}, \enskip p_{\rho_b}=a
    \end{cases},
    \label{eqn:second_deriative_lambda}
\end{gather}
$g(u,v,\mu,\nu)$ contains terms in \eqref{eq:old_newton_forward} and $\Xi(v)$ is defined analogously in terms of the appropriate $v$ and $\lambda$'s. Observe that $\Xi(u)$ and $\Xi(v)$ are diagonally dominant and symmetric, implying that they are positive definite. In the backward pass, we require the gradients of the loss $L$ with respect to $P$ and $\lambda$. Similar to prior work \cite{gould2016differentiating,amos2017optnet,ling2018game}, this may be done by applying the implicit function theorem or by simply manipulating differentials. This yields the gradients $ \nabla_P L = y_u v^T + u y_v^T$; $\nabla_{\lambda_h} L = \kappa_h ^T y_u$ for $h \in \mathcal{I}_u$, $\nabla_{\lambda_h} L = -K_h ^T y_v$ for $h \in \mathcal{I}_v$, where
\begin{gather}
    (\kappa_h)_a = 
    \begin{cases}
        1 + \log(u_a/u_{p_{\rho_a}}), \qquad& \rho_a = h \\
        - 1, \qquad& h \in C_a
    \end{cases} \\ 
    (K_h)_a = 
    \begin{cases}
        1 + \log(v_a/v_{p_{\rho_a}}), \qquad& \rho_a = h \\
        - 1, \qquad& h \in C_a
    \end{cases} \\
    \begin{bmatrix}
    y_u \\
    y_v \\
    y_{\mu} \\
    y_{\nu}
    \end{bmatrix} = 
    \begin{bmatrix}
    -\Xi(u) & P & E & 0 \\
    P^T & \Xi(v) & 0 & F \\
    E^T & 0 & 0 & 0 \\
    0 & F^T & 0 & 0
    \end{bmatrix}^{-1}
    \begin{bmatrix}
    -\nabla_u L \\
    -\nabla_v L \\
    0 \\
    0
    \end{bmatrix}. \label{eqn:backward_problem}
\end{gather}
\section{Fast forward and backward pass solvers}
In the framework of \citeauthor{ling2018game}, each gradient step in Algorithm~\ref{alg:lck_prev_work} involves solving an optimization problem. Thus, having efficient solvers is crucial in scaling up. In the naive solver, the forward pass is solved using Newton's method and we need to solve the system of linear equations (\ref{eqn:second_deriative_lambda}) in each iteration.
When the game tree is large, solving the system of linear equations in \eqref{eqn:second_deriative_lambda} multiple times dramatically slows down training.
Similarly for the backward pass, one needs to solve a single linear system shown in \eqref{eqn:backward_problem}. When the game is large, naively solving the linear system is also prohibitively slow, even when utilizing sparse matrices.
This serves as motivation for a first-order iterative method (FOM) which do not require the solution of a linear system as a subroutine. FOMs are also computationally attractive for solving extensive form games because of the underlying tree structures in games which may be exploited. We will focus on optimization problems in the following min-max form.
\begin{align}
    \min_{Ex=x_0} \max_{Fy=y_0} x^TPy + \mathcal{E}(x) - \mathcal{F}(y), 
    \label{eqn:chambolle}
\end{align}
where $\mathcal{E}(x)$ and $\mathcal{F}(y)$ are strictly convex functions. It is obvious from \eqref{eqn:seq_regularization_lambd} that the forward pass in our problem solves a problem in this form. We will show later that the backward pass problem shown in \eqref{eqn:backward_problem} can also be seen as solving a problem in this format. 

Many methods to solve \eqref{eqn:chambolle} have been proposed. In this paper, we adapt the method proposed by \citeauthor{chambolle2016ergodic} \shortcite{chambolle2016ergodic}\footnote{Note that the algorithm by \citeauthor{chambolle2016ergodic} is more general and applies beyond game solving.}. This, as well as many other first order methods apply best response subroutines towards smoothed versions of the min or max original problem taken in isolation. The solution is obtained by alternating between best-responses to minimization and maximization. Algorithm~\ref{alg:FOM_chambolle} gives the high-level overview of the optimization procedure, where BR are smoothed best responses with appropriately chosen Bregman divergences $D_x, D_y$ , their associated convex functions $\Psi_x, \Psi_y$, and `step sizes' $\tau, \alpha$.
\begin{align*}
    \text{BR}_x(\bar{x}, \tilde{y}) = \argmin_{Ex=x_0} x^T P \tilde{y} + \mathcal{E} (x) + \frac{1}{\tau} D_x(x, \bar{x}) \\
    \text{BR}_y(\bar{y}, \tilde{x}) = \argmin_{Fy=y_0} -\tilde{x}^T P y + \mathcal{F} (y) + \frac{1}{\sigma} D_y(y, \bar{y}).
\end{align*}
We first set $\tau=\sigma$ for convenience. For Algorithm~\ref{alg:FOM_chambolle} to be practical, we will require the best response oracles $\text{BR}_x, \text{BR}_y$ to be computed efficiently. By setting $D_x(x, \bar{x})$ and  $D_y(y, \bar{y})$ to be of specific form similar to $\mathcal{E}(x)$ and $ \mathcal{F}(y)$ respectively will simplify $\text{BR}_x$ and $\text{BR}_y$ to be (up to a factor) of the forms 
\begin{align*}
    \argmin_{Ex=x_0} x^T c_x + \mathcal{E}(x) \quad \text{and} \quad
    \argmin_{Fy=y_0} y^T c_y + \mathcal{F}(y),
\end{align*}
which are efficiently computed by exploiting the structure of the problem in extensive form games. This avoids the need to solve a linear system with $P$ as part of the design matrix. 
The remainder of this section outlines the procedures required for both forward and backward passes.
For brevity, we discuss this from the view of the minimization -- the maximization subroutine is entirely analogous. Lastly, we remark that the computational advances in this section are independent of the NLQRE, i.e., they remains applicable to the original framework of \citeauthor{ling2018game}.
\begin{algorithm}
\SetAlgoLined
 \KwIn{$x^{(0)}, y^{(0)}, P, \tau, \sigma$}
 \For{$i$ in $\{0, \dots \}$}
 {
  $\tilde{y}=y^{(i)}$\;
  $x^{(i+1)} = \text{BR}_x(x^{(i)}, \tilde{y}; P, \tau)$\;
  $\tilde{x} = 2x^{(i+1)} - x^{(i)}$\;
  $y^{(i+1)} = \text{BR}_y(y^{(i)}, \tilde{x}; P, \sigma)$\;
 }
 \caption{FOM method to solve \eqref{eqn:chambolle}}
 \label{alg:FOM_chambolle}
\end{algorithm}
\subsubsection{Forward Pass}
For this section, the $u=x, v=y$ when referring Algorithm~\ref{alg:FOM_chambolle}. Setting $\mathcal{E}(u), \mathcal{F}(v)$ to be the entropy terms in \eqref{eqn:seq_regularization_lambd} and $u_0 = e, v_0=f$ gives the expression in the form of \eqref{eqn:chambolle}. The natural divergence to be chosen is the standard entropy divergence adapted to the dilated setting (dropping terms in $D_u$ which do not contain $u$).
\begin{align*}
    \Psi_u(u) &= \mathcal{E}(u) = \sum_{h \in \mathcal{I}_u} \lambda_h \sum_{a \in \mathcal{A}_h} u_a \log \frac{u_a}{u_{p_h}}\\
    D_u(u, \bar{u}) &=  \Psi_u(u) - u^T  \Psi_u'(\bar{u}) \\
    \Psi_u'(\bar{u}) &= \lambda_{\rho_a} - \sum_{h' \in C_a} \lambda_{h'} + \lambda_{\rho_a} \log \frac{\bar{u}_a}{\bar{u}_{p_a}}
\end{align*}
where a similar expression holds for $D_v(v, \bar{v})$. Plugging into the expression for $\text{BR}_u$ gives 
\begin{align*}
    \text{BR}_u(\bar{u}, \tilde{v}) = \argmin_{Eu=e} \frac{\tau}{1+\tau} u^T \left( P \tilde{v} + \Psi_u'(\bar{u}) \right) + \mathcal{E}(u).
\end{align*}
It is known that, $\text{BR}_u(\bar{u}, \tilde{v})$ may be solved by a single bottom-up traversal of the game tree and a single sparse matrix-vector multiplication \cite{hoda2010smoothing}. At each information set , we solve for the `behavioral' best response (i.e., assuming that information set was the root). Each of these sub-problems may be expressed in closed form using log-sum-exp and softmax functions. The sequence form is recovered from behavioral strategies with a single downwards traversal of the tree. The precise details are contained in the appendix.

\subsubsection{Backward Pass}
The backward pass also requires solving a linear system to obtain $[y_u \enskip y_v \enskip y_\mu \enskip y_\nu]$. We first begin by making the crucial observation that the (necessary and sufficient) KKT conditions of the following optimization problem is precisely the linear system in \eqref{eqn:backward_problem}. 
\begin{align}
\label{eq:kkt_equiv}
    \min_x \max_y &\quad 
    x^T P y 
    + \frac{1}{2} x^T \Xi(u) x
    - \frac{1}{2} y^T \Xi(v) y \nonumber 
    \\& \qquad + \nabla_u L^T x  
    + \nabla_v L^T y \nonumber \\
    \text{subject to} &\quad E x = 0 \qquad 
    F y = 0.
\end{align}
Note that $u$ and $v$ are constants in the backwards pass, here we are optimizing over $x, y$, which are  \textit{not} probabilities. Since $\Xi(u)$ and $\Xi(v)$ are positive definite, this is a convex-concave problem of the form required by Algorithm~\ref{alg:FOM_chambolle}. We select the natural distance generating function $\Psi_x = \frac{1}{2} x^T \Xi(u) x$ which yields (ignoring terms containing only $\bar{x}$),
\begin{align*}
    D_x(x, \bar{x}) &= \frac{1}{2} x^T \Xi(u) x - x^T \Xi(u) \bar{x}
\end{align*}
Plugging this into the expression for $\text{BR}_x(\bar{x}, \tilde{y})$ and rearranging gives
\begin{align}
    \argmin_{Ex=0} \frac{\tau}{1+\tau} x^T (\nabla_u L + P\tilde{y} - \frac{1}{\tau}\Xi(u) \bar{x}) +  \frac{1}{2} x^T \Xi(u) x
    \label{eq:backward_optimization}
\end{align}
Letting $c=\frac{\tau}{1+\tau} \left(\nabla_uL + P\tilde{y} - \frac{1}{\tau}\Xi(u)\bar{x} \right)$ in \eqref{eq:backward_optimization} gives the KKT conditions 
\begin{align*}
    c + E^T \gamma + \Xi(u) x &= 0, \qquad Ex = 0
\end{align*}
where $\gamma$ are Lagrange multipliers. Multiplying by $E\Xi^{-1}(u)$ gives a linear system in $\gamma$
\begin{align}
    E \Xi^{-1}(u) c + E\Xi^{-1}E^T \gamma &= 0.
    \label{eq:solve_gamma}
\end{align}
After solving for $\gamma$, one may solve for $x$
\begin{align}
    x &= \Xi^{-1}(u) \left(  -c - E^T \gamma \right).
    \label{eq:solve_x}
\end{align}
\begin{proposition}
Solving for $\gamma$ and $x$ in Equations \eqref{eq:solve_gamma} and \eqref{eq:solve_x} require linear time (in the size of the game tree).
\label{thm:main_speed}
\end{proposition}
The derivation involves exploiting the tree-structure inherent in extensive form games. Computational details and proofs are deferred to the appendix.
\section{Experiments}
\label{sec:expts}
The proposed first order method was implemented using Cython. We chose to do so since the best-response subroutines require tree-traversals, which are expensive in Python. while the second order method used the Numpy and Scipy libraries for the solution of linear systems. Where possible, we utilized the Scipy sparse matrix library. This was seen to provide a significant speedup for sparse $P$ for both our method and Newton's method. The PyTorch automatic differentiation library \cite{paszke2017automatic} was used to automatically obtain gradients for components outside the game solving module.
\subsection{Synthetic datasets}
Here we use randomly generated extensive form games to illustrate the computational efficiency of our proposed first order method compared to the second order method used by \citeauthor{ling2018game}. We evaluate the solvers for the forward and backward passes in isolation. The experiments are run over several depths $d$. Normal form games have $d=1$. When $d>1$, we adopt the following extensive form game. players play $d$ distinct simultaneous sub-games in succession, where each simultaneous sub-game has $\hat{n}$ actions. Transitions to the next sub-game is governed by the joint action by both players, i.e., the size of $P$ will be exponential in $d$. The payoff matrices $P$ were generated with each non-zero entry uniformly chosen from $[-10, 10]$, and rationality parameter $\lambda$ for \textit{each} information set uniformly and independently chosen between $[0.9, 1.1]$. All timings presented are wall-clock timings. Experiments are run on the cloud with identical Amazon EC2 instances. We set $\tau=0.1$ for all evaluations.
\subsubsection{Evaluation of forward passes}

In the forward pass, we compared the baseline Newton solver to our proposed first-order method. However, the termination criterion for the 2 methods are non-identical; as Newton's method minimizes the residual rather than duality gap. To strike a fair comparison, we evaluated the 2 methods by first running Newton's method till a residual of less than $10^{-3}$ is achieved. The duality gap of that solution is computed and subsequently used as the termination criterion for the FOM\footnote{On occasion, the Newton solver gave a gap extremely close to numerical precision. In these cases, we apply to a termination criterion of $10^{-12}$.}. The timings and speedup are averaged over $50$ trials and presented in Figure~\ref{fig:synth_forward_pass}.
\begin{figure}
    \begin{subfigure}[b]{0.48\linewidth}
    \centering
    \includegraphics[width=\textwidth]{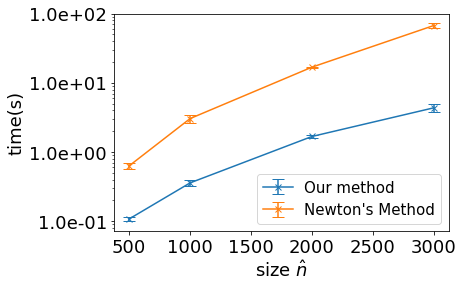}
    \end{subfigure}
    \begin{subfigure}[b]{0.48\linewidth}
    \centering
    \includegraphics[width=\textwidth]{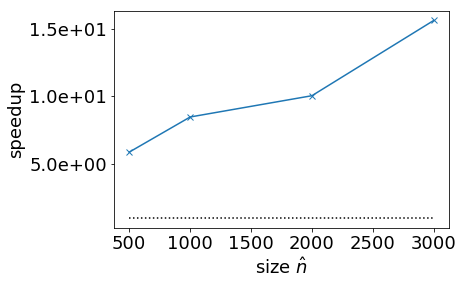}
    \end{subfigure}
    \\
    \begin{subfigure}[b]{0.48\linewidth}
    \centering
    \includegraphics[width=\textwidth]{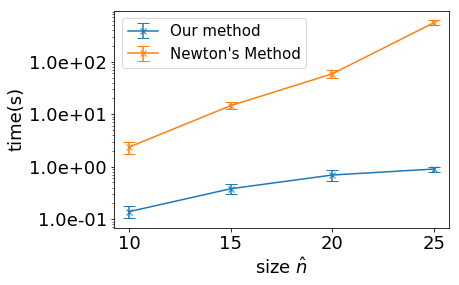}
    \end{subfigure}
    \begin{subfigure}[b]{0.48\linewidth}
    \centering
    \includegraphics[width=\textwidth]{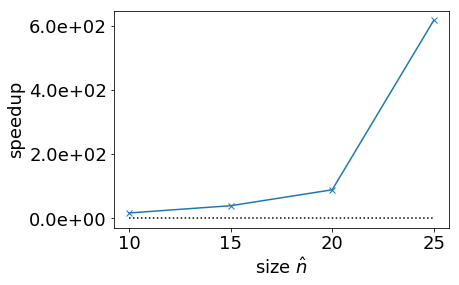}
    \end{subfigure}
    \label{fig:synth_forward_pass}
    \caption{Timings (left) and speedup (right) for forward passes. From top to bottom, $d=1,2$. Error bars represent 1 standard deviation. Dotted lines are optimal results given the ground truth.}
    \label{fig:synth_forward_pass}
\end{figure}
\subsubsection{Evaluation of backward passes}
\begin{figure}[ht]
    \begin{subfigure}[b]{0.48\linewidth}
    \centering
    \includegraphics[width=\textwidth]{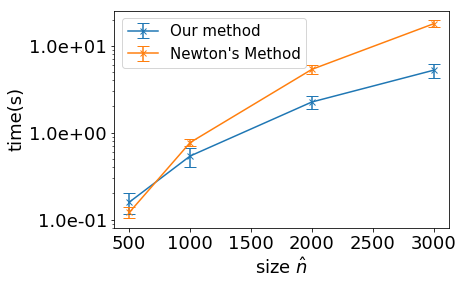}
    \end{subfigure}
    \begin{subfigure}[b]{0.47\linewidth}
    \centering
    \includegraphics[width=\textwidth]{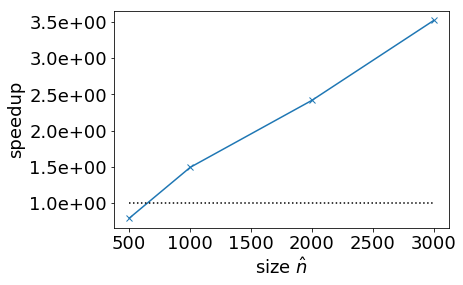}
    \end{subfigure}
    \\
    \begin{subfigure}[b]{0.48\linewidth}
    \centering
    \includegraphics[width=\textwidth]{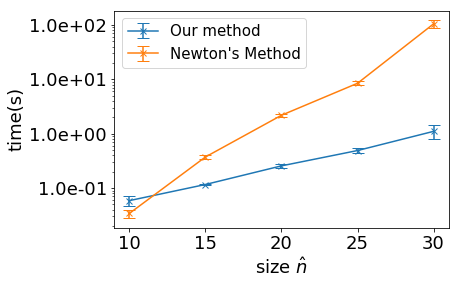}
    \end{subfigure}
    \begin{subfigure}[b]{0.48\linewidth}
    \centering
    \includegraphics[width=\textwidth]{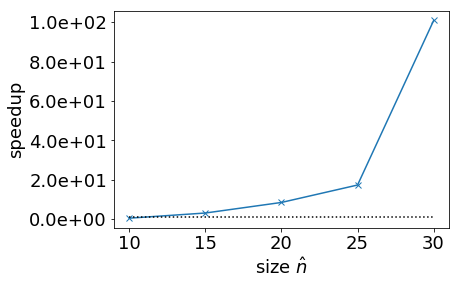}
    \end{subfigure}\\
    \begin{subfigure}[b]{0.48\linewidth}
    \centering
    \includegraphics[width=\textwidth]{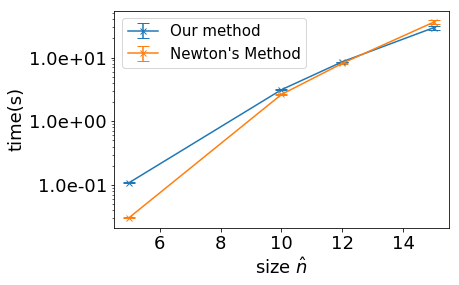}
    \end{subfigure}
    \begin{subfigure}[b]{0.48\linewidth}
    \centering
    \includegraphics[width=\textwidth]{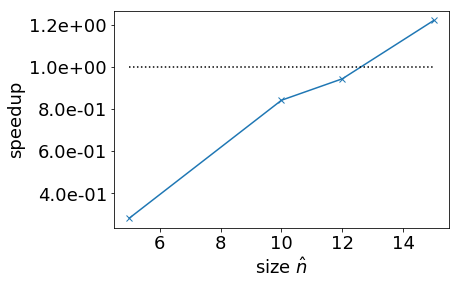}
    \end{subfigure}
    \caption{Timings (left) and speedup (right) for backward passes. From top to bottom, $d=1,2,3$. Error bars represent 1 standard deviation. Dotted lines are optimal results given the ground truth.}
    \label{fig:synth_backward_pass}
\end{figure}

In the backward pass, the comparison for our proposed FOM is against solving the linear system in \eqref{eqn:backward_problem} directly. In the loss function, we will concern ourselves with the setting where the true matrix $P$ and $\lambda$ parameters are used in computing $u, v$ for the forward pass. 
This corresponds to the case the model is already fairly well trained. The results over 50 trials are presented in Figure~\ref{fig:synth_backward_pass}.
It is clear from both figures that our method scales much better than Newton's method for randomly generated matrices. Speedups of more than an order of magnitude are fairly common, and the improvement increases with problem size. Furthermore, it was also observed that our method consumed far less memory than sparse solvers. In fact, solving the sparse system when $d=3, \hat{n}=17$ (not plotted) required more than $10$GB of memory. On the other hand, our FOM was able to solve such instances in less than a minute, and with no noticeable increase in memory usage. Note that $P$ contains more than 1.4 million rows and columns in this setting.
\subsection{One-Card Poker}
Here we evaluate our method on the game of one card poker. This multi-card extension of Kuhn poker contains most interesting strategic elements of game playing (e.g., bluffing) and was used by \citeauthor{ling2018game} to illustrate that it is possible to learn distribution of cards in a deck just by observation player actions. However, the authors worked only on tiny settings with just $4$ cards -- in sequence form, player strategies may be represented in a $16$ dimension vector. Furthermore, the authors assumed that there were no varying input features (i.e., the card distributions were identical for each action observed). These assumptions enabled them to achieve significant speedup by solving the game just once in the forward pass, rather than once for each point in the mini-batch. As we will see, their solver is too slow to be of practical use in larger or featurized settings.

Here, we operate in a slightly different setting. Instead of trying to learn underlying card distributions, we learn player rationality parameters. We assume that player rationality is independent of the cards being drawn, and only depends on the past actions of (both) players. In this setting, there are just $4$ parameters to be learned. This is independent of the size of the deck. 

We generate our data assuming that player rationality is some linear function of a scalar feature, i.e., there are $4$ weights to be learned. The weight vector is drawn uniformly from $[0, 0.01]$. Feature vectors are drawn between $[0, 1]$. Our model is $\lambda_h = w_i \times f + \epsilon_{\lambda}$. The addition of a small $\epsilon_{\lambda}$ ensures that the $\lambda$'s will always remain positive; in our experiments, $\epsilon_{\lambda}=0.001$. For each feature, we compute the $\lambda$'s and find its corresponding equilibrium from which we sample player actions. The training set of size $2000$, with an independent test set of size $1000$. We minimized the log loss using the Adam optimizer with a batch size of $64$ and learning rate of $10^{-4}$.
\begin{figure}
    \begin{subfigure}[b]{0.47\linewidth}
    \centering
    \includegraphics[width=\textwidth]{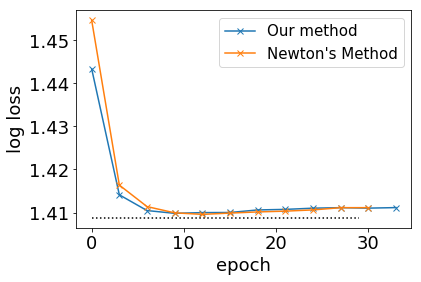}
    \end{subfigure}
    \begin{subfigure}[b]{0.46\linewidth}
    \centering
    \includegraphics[width=\textwidth]{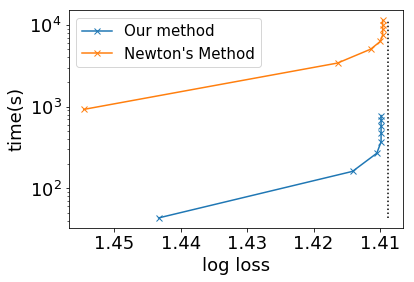}
    \end{subfigure}
    \\
    \begin{subfigure}[b]{0.47\linewidth}
    \centering
    \includegraphics[width=\textwidth]{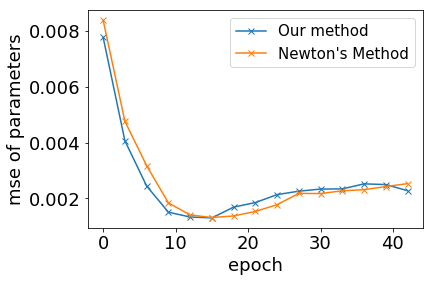}
    \end{subfigure}
    \begin{subfigure}[b]{0.46\linewidth}
    \centering
    \includegraphics[width=\textwidth]{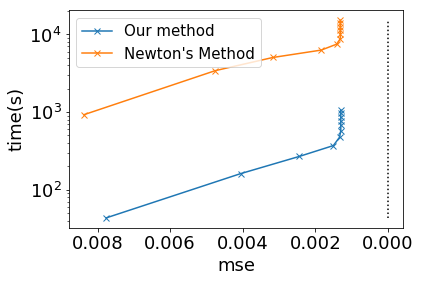}
    \end{subfigure}
    \caption{Left: log-loss and mse per epoch, Right: time required to obtain a particular log loss or mse. Dotted lines are optimal results given the ground truth.}
    \label{fig:poker_results}
\end{figure}

We compared our solver against Newton's method, which terminates at a residual of $10^{-8}$. We fixed $\tau=1$ for the forward solver and $\tau=0.1$ for the backward solver. 
The results are plotted in Figure~\ref{fig:poker_results}.

In all $3$ cases, both the log-loss is close to optimal given around $30$ epochs.  As expected, our exacts solver exhibits behavior almost identical to that of Newton's method on a per-epoch basis. 
Our solver is significantly faster than the baseline. It was observed that at almost all stages of training, Newton's method took almost $2$ orders of magnitude time in order to learn a model of similar performance. In fact, a single epoch using Newton's method takes as much time as training the entire model using our solver. 


\subsection{Information gathering dataset}
Here we demonstrate the applicability of the nested logit model (i.e., a one player game) using a publically available dataset \cite{hunt2016approach}. The game proceeds as follows. Suppose there are 4 faced-down cards ranging from 1-10 placed in a $2 \times 2$ matrix (with potential repetitions). The goal of the game is to select the row containing cards with the largest sum. The game proceeds in 4 stages. At each stage of the game, the player may make a guess prematurely, or spend some points in revealing a new card. At the fourth and final stage, the player has to make a guess. The player obtains a reward of 60 and -50 points for correct and incorrect guesses, and may only guess once. The challenge is for the player to judge if it is worth paying to gather more information. Computationally, the optimal policy may be easily obtained using dynamic programming. 

However, humans are rarely perfectly rational. We model bounded rationality using the nested logit model. It is assumed that the level of rationality should be a function of a) how many cards are already open, and b) side information such as one's educational qualifications. This leads to a natural description of the game with $4$ different $\lambda$'s, each of which is some function of features, which we describe below. 

\begin{figure}[ht]
    \begin{subfigure}[b]{0.47\linewidth}
    \centering
    
    \includegraphics[width=1.0\textwidth]{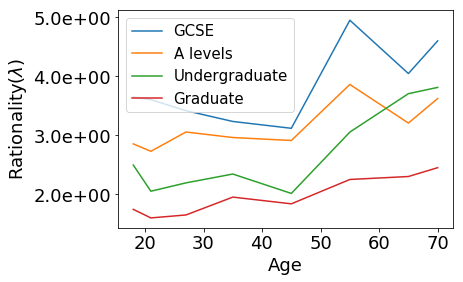}
    \label{fig:rat1}
    \end{subfigure}
    \begin{subfigure}[b]{0.47\linewidth}
    \centering
    \includegraphics[width=1.0\textwidth]{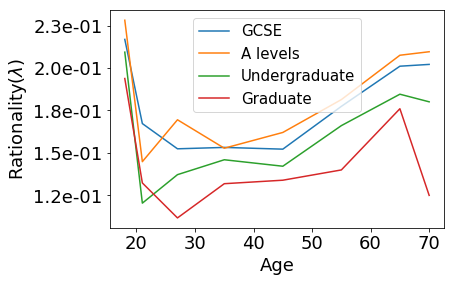}
    \label{fig:rat2}
    \end{subfigure}
    \\
    \begin{subfigure}[b]{0.47\linewidth}
    \centering
    \includegraphics[width=1.0\textwidth]{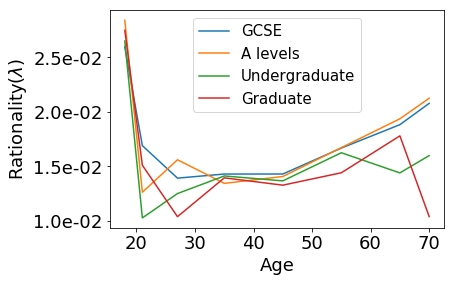}
    \label{fig:rat3}
    \end{subfigure}
    \begin{subfigure}[b]{0.47\linewidth}
    \centering
    \includegraphics[width=1.0\textwidth]{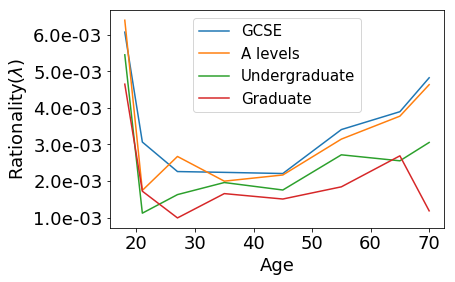}
    \label{fig:rat4}
    \end{subfigure}
    \caption{Rationality parameters $\lambda$ for each stage of the game as a function of age and education. Stages are presented in row-major ordering. Note that ages are binned within some range. All respondents below the age of 18 were grouped together in a single age range.} 
    \label{fig:hunt_lambdas}
\end{figure}

Two models are trained for this experiment. \textsc{NoFeat} refers to the case when there are no features (i.e., we are simply learning $\lambda$) and \textsc{Feat} when we are exploiting demographic information. In this case, features comprise the player's academic qualifications and age, both with one-hot encodings. A player's age is split into 8 age ranges, and education levels follow that of the UK (i.e., GCSE, A levels, Undergraduate, Graduate). Our model employs a neural network with 3 hidden layers of width 100, interspersed by rectified linear activations. To ensure $\lambda$'s are positive, all inputs were exponentiated before being fed into the solver. Figure~\ref{fig:hunt_losses} shows the log loss over the overall game as well as the loss at each individual stage. For comparison, we also provide the results for a player who picks a random action at every stage of the game. The learned $\lambda$ parameters for each configuration of features is presented in Figure~\ref{fig:hunt_lambdas}.
\begin{figure}
    \centering
    \begin{tabular}{l*{6}{c@{\hspace{1.5\tabcolsep}}}r}
              & loss & loss(1) & loss(2) & loss(3) & loss(4) \\
    \hline
    \textsc{Uniform} & 1.833 & 1.099 & 1.099 & 1.099 & 0.693 \\
    \textsc{NoFeat} & 1.422 & 0.878 & \textbf{0.863} & 0.826 & \textbf{0.130}\\
    \textsc{Feat} &  \textbf{1.419}       & \textbf{0.874} & 0.866 & \textbf{0.818} & 0.145  \\
    \end{tabular}%
    \caption{Log losses for the information gathering dataset. The first column shows losses over the whole game, other columns show losses for individual stages.}
    \label{fig:hunt_losses}
\end{figure}
From Figure~\ref{fig:hunt_losses}, we can see that both trained models significantly outperform \textsc{Uniform}. Log losses at each stage appear to decrease with the stage number. This is unsurprising since players behave more rationally (and hence predictably) as more information is revealed. However, our model appears to perform worse at stage $4$, which is in fact a problem with full observability. We suspect this higher loss is a consequence of our model `overfitting' to be overly confident at the final stage, incurring a huge loss on the rare occasion a player answers incorrectly. 

Several trends are observed from Figure~\ref{fig:hunt_lambdas}. First, notice that $\lambda$ decrease by approximately a factor of $10$ between stages. This is fairly expected, since each information gathering leads to $10$ other potential states. Also unsurprisingly, better educated respondents exhibit more rationality (recall a lower $\lambda$ implies a more rational player). Interestingly, we can see a U-shaped trend in all stages, suggesting that people in the mid twenties and thirties are most rational. Both these observations agree with the findings by \cite{hunt2016approach}, where it was shown that higher educated and middle aged respondents obtained the most reward.
\section{Conclusion}
In this paper, we substantially improve upon existing work in differentiable game learning. We propose the NLQRE which generalizes QRE and NL models. We also derive gradients for backpropagation and learning, and develop solvers which lead to speedups of several orders of magnitude. Future work include the learning of game structure and extensions for general-sum games.
\bibliography{aaai19}
\bibliographystyle{aaai}
\clearpage
\newpage
\section*{Appendix}
\subsection*{Best response of the NLQRE as a nested logit}
Here we show that the best-response in the NLQRE, which may contain parallel information sets (either due to chance or actions by other players) may be regarded as a nested logit. The idea is to express each strategy in the reduced normal form into a sequence of decisions, each describing what is to be done in each parallel information set.

\begin{figure}[ht]
    \centering
    \includegraphics[width=0.75\linewidth]{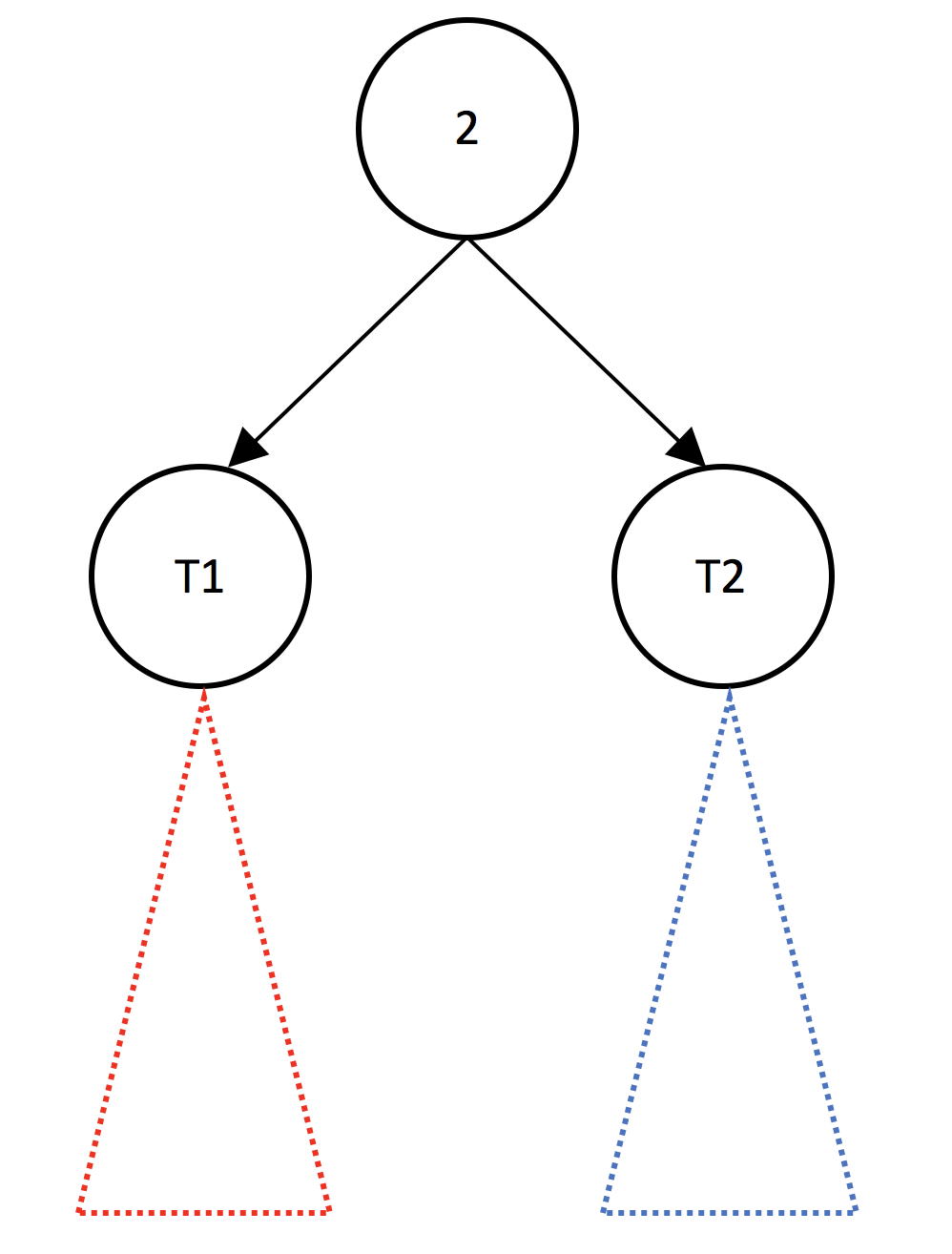}
    \caption{Full 2-player game tree. The first action is performed by chance (or the other player). The player has to provide for contingencies in both the red and blue trees.}
    \label{fig:full_gametree}
\end{figure}

Consider the following game in Figure~\ref{fig:full_gametree}. Chance (or the other player), labelled as (2), first chooses out of $2$ actions, which is made known to the player. For example, this could be the private cards which are dealt to a player in a game of poker. The nodes $T_1$ and $T_2$ are the subtrees following these $2$ actions by the chance player.

\begin{figure}[ht]
    \centering
    \includegraphics[width=0.75\linewidth]{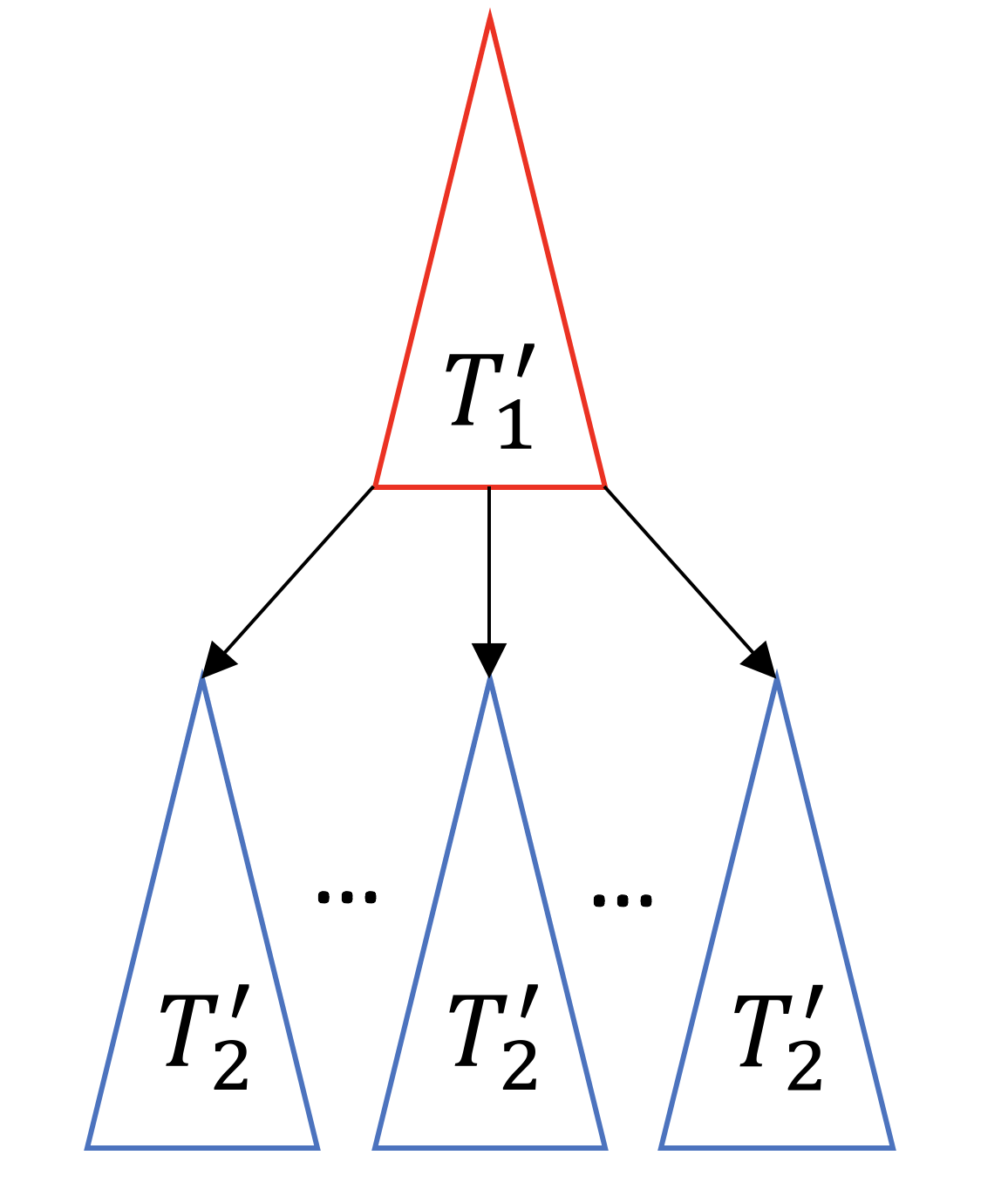}
    \caption{Player's decision tree. Red and blue components are decision trees for the player. Each leaf (or path from the root to a leaf) is a pure strategy in the reduced normal form of the game in Figure~\ref{fig:full_gametree}.}
    \label{fig:stacked_gametree}
\end{figure}

Without loss of generality, let $T_1'$ and $T_2'$ be the tree representation of the player's strategies in subgame $T_1$ and $T_2$, i.e., $T_1'$ and $T_2'$ are decision trees for the player. Given that chance could have chosen either action to begin with, the pure strategies are the cross product of all strategies between $T_1'$ and $T_2'$, i.e., the player has to account for all possible contingencies. This may be written as a 2-stage decision process in Figure~\ref{fig:stacked_gametree}, where the first and second stages are choices from $T_1'$ and $T_2'$ respectively, where $T_2'$ is duplicated $n_{\text{leaves}}$ times, where $n_{\text{leaves}}$ is the number of possible leaves for $T_1'$. 

The rewards are additive in each stage, implying that the best response for each of the duplicated trees are identical. (Note however that the actual payoffs are modulated by the probability of the chance player choosing the left or right action, but this factor is identical for each  copy of $T_2'$). Furthermore, the leaves of $T_1'$ form a probability vector (since $T_1'$ is a decision tree). This implies that setting the rationality parameters for the roots of all copies $T_2'$ to be $\lambda_2$ and the rationality parameter at $T_1'$ to be $\lambda_1$ yields precisely one-player version of the optimization problem in \eqref{eqn:seq_regularization_lambd}, since the objective in each copy of $T_2'$ is identical and their coefficients sum to $1$. Recursively applying this process bottom up to each subtree (i.e., making duplicate copies of subtrees whenever we encounter parallel information sets) gives the desired result.

Observe that each pure strategy (path) in Figure~\ref{fig:stacked_gametree} is a pure strategy in reduced normal form. However, each path may pass through different information sets (for example, when there is nesting of actions in the bus example), and hence different $\lambda$ parameters. This is in line with what one would expect with nested logits. 

\subsection*{Fast forward and backward pass solvers}
In this section, we provide the complete computational details and proofs with regard of how to compute best responses for the forward and backward passes. 
\subsubsection{Forward Pass}
For this section, the $u=x, v=y$ when referring to Algorithm~\ref{alg:FOM_chambolle}. Setting $\mathcal{E}(u), \mathcal{F}(v)$ to be the entropy terms in \eqref{eqn:seq_regularization_lambd} and $u_0 = e, v_0=f$ gives the expression in the form of \eqref{eqn:chambolle}. The natural divergence to be chosen is the standard entropy divergence adapted to the dilated setting (dropping terms in $D_u$ which do not contain $u$).
\begin{align*}
    \Psi_u(u) &= \mathcal{E}(u) = \sum_{h \in \mathcal{I}_u} \lambda_h \sum_{a \in \mathcal{A}_h} u_a \log \frac{u_a}{u_{p_h}}\\
    D_u(u, \bar{u}) &=  \Psi_u(u) - u^T  \Psi_u'(\bar{u}) \\
    \Psi_u'(\bar{u}) &= \lambda_{\rho_a} - \sum_{h' \in C_a} \lambda_{h'} + \lambda_{\rho_a} \log \frac{\bar{u}_a}{\bar{u}_{p_a}}
\end{align*}
where a similar expression holds for $D_v(v, \bar{v})$. Plugging into the expression for $\text{BR}_u$ gives 
\begin{align*}
    \text{BR}_u(\bar{u}, \tilde{v}) = \argmin_{Eu=e} \frac{\tau}{1+\tau} u^T \left( P \tilde{v} + \Psi_u'(\bar{u}) \right) + \mathcal{E}(u).
\end{align*}
It is known that, $\text{BR}_u(\bar{u}, \tilde{v})$ may be solved by a single bottom-up traversal of the game tree and a single sparse matrix-vector multiplication \cite{hoda2010smoothing}. At each information set , we solve for the `behavioral' best response (i.e., assuming that information set was the root). Each of these sub-problems may be conveniently expressed using log-sum-exp and softmax functions. Denoting $c_u=P \tilde{v} + \Psi_u'(\bar{u})$, we compute
\begin{align*}
    \min_{u, Eu=e} u^Tc_u + \sum_{h \in \mathcal{I}_u} \lambda_h u_{p_h} \sum_{a \in \mathcal{A}_h} \frac{u_a}{u_{p_h}} \log \left( \frac{u_a}{u_{p_h}} \right),
\end{align*}
where the constraint that $u > 0 $ is implicit from the log barrier.

Optimization of the inner summation, along with the relevant part of the inner product may be done in closed form using log-sum-exp. The tree constraints for $u$ allows them to perform traversals bottom up. Throughout the traversal process, denote $z_h$  as the `value' of each infoset $h$ and $r_a$ as the value of each action.
\begin{gather*}
    z_h = \lambda_h \log \left( \sum_{\rho_a = i} \exp \left( \frac{r_a}{\lambda_h} \right) \right), \quad
    r_a = -c_a + \sum_{h \in C(a)} z_h
\end{gather*}
The behavioral strategies $\hat{u}_a$ may be expressed using the softmax function. For an action $a$ belonging to info set $h$, 
\begin{align}
    \hat{u}_a &= \frac{\exp(r_a/\lambda_h)}{\sum_{a' \in C_h} \exp(r_{a'}/\lambda_h)}
\end{align}
The sequence form may be recovered from behavioral strategies using a single downwards traversal of the tree.

\subsubsection{Backward Pass}
The backward pass also requires solving a linear system to obtain $[y_u \enskip y_v \enskip y_\mu \enskip y_\nu]$. By rewriting the linear system as another min-max problem, we may again apply Algorithm~\ref{alg:FOM_chambolle}. Observe that the solution to the system are precisely the (necessary and sufficient) KKT conditions of the following min-max problem
\begin{align}
\label{eq:kkt_equiv}
    \min_x \max_y &\quad 
    x^T P y 
    + \frac{1}{2} x^T \Xi(u) x
    - \frac{1}{2} y^T \Xi(v) y \nonumber 
    \\& \qquad + \nabla_u L^T x  
    + \nabla_v L^T y \nonumber \\
    \text{subject to} &\quad E x = 0 \qquad 
    F y = 0.
\end{align}
Note that $u$ and $v$ are constants in the backwards pass, here we are optimizing over $x, y$, which are  \textit{not} probabilities. Since $\Xi(u)$ and $\Xi(v)$ are positive definite, this is of the form required by Algorithm~\ref{alg:FOM_chambolle}. We select the natural distance generating function $\Psi_x = \frac{1}{2} x^T \Xi(u) x$ which yields (ignoring terms containing only $\bar{x}$),
\begin{align*}
    D_x(x, \bar{x}) &= \frac{1}{2} x^T \Xi(u) x - x^T \Xi(u) \bar{x}
\end{align*}
Plugging this into the expression for $\text{BR}_x(\bar{x}, \tilde{y})$ and rearranging gives
\begin{align}
    \argmin_{Ex=0} \frac{\tau}{1+\tau} x^T (\nabla_u L + P\tilde{y} - \frac{1}{\tau}\Xi(u) \bar{x}) +  \frac{1}{2} x^T \Xi(u) x
    \label{eq:backward_optimization}
\end{align}
\paragraph{Proposition~\ref{thm:main_speed}} \textit{
Solving for $\gamma$ and $x$ in Equations \eqref{eq:solve_gamma} and \eqref{eq:solve_x} require linear time (in the size of the game tree).}

\begin{proof}
To obtain efficient solutions for the best responses, we require the following handy results, which may all be verified by algebraic manipulation.
\begin{lemma}
\label{thm:Xi_inv_ET}
$\Xi^{-1}(u) E^T = \begin{bmatrix} \alpha_1, \alpha_2, \dots ,\alpha_{|\mathcal{I}_u|} \end{bmatrix}$, where $\alpha_i$'s are column vectors of size equal to number of actions, and contain $u_a/\lambda_h$'s where $a$ is some descendent of information set $i$, and 0 otherwise.
\end{lemma}
Taking transposes gives the following, $E \Xi^{-1}(u)$ is equal to $\begin{bmatrix} \beta_1, \dots, \beta_{|\mathcal{A}_u|}\end{bmatrix}$, where the $\beta_a$'s are column vectors of length equal to $|I_u|$, and have entries equal to $\frac{u_a}{\lambda_h}$ if the index $i$ is an ancestor of $a$.
\begin{lemma}
\label{thm:E_Xi_inv_ET}
$E\Xi^{-1}(u) E^T = \text{diag}(\{u_{p_h}\})$, i.e. equal to a square matrix of size $|I_u|$, with diagonal entries equal to the $\frac{u_{p_h}}{\lambda_{h}}$ corresponding to a given information state's parent action.
\end{lemma}
\begin{lemma}
\label{thm:E_Xi_inv_c_FAST}
For any vector $c$, $E\Xi^{-1}(u)c$ may be computed in linear time by traversing the tree bottom-up.
\end{lemma}
We are now ready to prove the main theorem. Letting $c=\frac{\tau}{1+\tau} \left(\nabla_uL + P\tilde{y} - \frac{1}{\tau}\Xi(u)\bar{x} \right)$ in \eqref{eq:backward_optimization} gives
\begin{align}
    \text{BR}_x(\bar{x},\tilde{y}) &= \min_{x, Ex=0} x^T c + \frac{1}{2} x^T \Xi(u) x,
\end{align}
which has KKT conditions 
\begin{align}
    c + E^T \gamma + \Xi(u) x &= 0 \\
    Ex &= 0
\end{align}
Multiplying by $E\Xi^{-1}(u)$ gives
\begin{align}
    E \Xi^{-1}(u) c + E\Xi^{-1}E^T \gamma &= 0
\end{align}

Note that we should not have introduced new roots in doing so, since these are linear systems and there is a unique solution to $\gamma$ both before and after the multiplication. Applying Lemma~\ref{thm:Xi_inv_ET} and Lemma~\ref{thm:E_Xi_inv_ET} gives an expression for $\gamma$. Lemma~\ref{thm:E_Xi_inv_c_FAST}, together with the fact that $E\Xi^{-1}E^T$ is diagonal implies that $\gamma$ may be solved for in linear time. The computation of $c$ requires $Py$, which may be done in time linear in the size of the \textit{extensive form} game tree. In the extreme case, the game could be a single-stage simultaneous move game, resulting in $P$ being dense. However, for typical EFGs, $P$ should be fairly sparse.

With $\gamma$, we may solve for $x$
\begin{align}
    x &= \Xi^{-1}(u) \left(  -c - E^T \gamma \right)
\end{align}
Since $\Xi(u)$ is tree-structured, the inversion may be done in linear time using Gaussian Elimination. Similarly, $E^T\gamma$ may be computed in linear time because of sparsity in $E$. That is, the number of non-zero elements in $E$ is equal to the sum of the number of actions over all information sets (recall that each row in $E$ has non-zero entries for the actions for a given information set and its parent).
\end{proof}

\end{document}